# A Heuristic Bayesian Approach to Knowledge Acquisition: Application to Analysis of Tissue-Type Plasminogen Activator

Ross D. Shachter[†], David M. Eddy, Vic Hasselblad, and Robert Wolpert
Center for Health Policy Research and Education
Duke University, PO Box GM, Durham, NC 27706

This paper describes a heuristic Bayesian method for computing probability distributions from experimental data, based upon the normal distribution form of the influence diagram. An example illustrates its use in medical technology assessment. This approach facilitates the integration of results from different studies, and permits a medical expert to make proper assessments without considerable statistical training.

There has been extensive research on the construction and manipulation of expert systems using probabilities as a measure for uncertainty. These systems are capable of recognizing considerable dependence and of learning from unreliable observations. These models are inherently *subjective*, and the probabilities on which they are based are the beliefs of some *decision maker*. As the evidence accumulates, the decision maker has less uncertainty about the states of the world and he can make a better decision. This framework has proven successful for learning about a given individual or situation. For example, a physician could incorporate her observations from tests and examinations into a distribution which describes her beliefs about a patient's underlying disorder. The subjectivity of the probabilities is a necessary "evil," since the responsibility for the decision clearly lies with the physician. In exchange, she has a powerful framework at her disposal.

In this paper, our goal is to develop the prior probabilities for the population parameters on which the individual model is based. We use the results of scientific experiments to learn about an entire class of situations or individuals, rather than just one. For example, we might derive the probability distribution over health outcomes for a class of patients undergoing treatment for a given health problem. As in the paradigm above, we revise our initial opinions on the basis of observations. There is still a need for subjective judgment, in weighing and sifting the experimental evidence. However, it is important that our results be defensible, since the probabilities (and conclusions) will be used by other decision makers in the provision of patient care and the selection of treatments. Therefore, our goal is to incorporate all relevant experimental data in our analysis, and to use it in an appropriate manner.

---

[†]Visiting CHPRE from the Department of Engineering-Economic Systems, Stanford University.

There are several technical problems which must be overcome in our work. First, there are often a large number of dependent variables, the population parameters. Second, since most of these variables are probabilities, they are necessarily continuous-valued. Finally, it takes specialized skills and expertise to properly interpret and integrate the experimental data so that it can be prepared for analysis. Therefore, we would like to have a methodology that requires minimal additional statistical knowledge to perform that analysis.

We have developed a methodology for solving this problem in order to analyze alternative medical technologies (Eddy, 1986). It has been used to solve many problems, including an analysis of the efficacy of tissue-type Plasminogen Activator (t-PA) in the treatment of heart attacks (Eddy, 1987). This analysis was performed using a prototype system (Hasselblad et al, 1986) written in APL on the PC. The system maintains probability distributions for each of the population parameters and manipulates them through one-dimensional numerical integration.

In this paper, we develop a heuristic approach to this problem, based on a multivariate normal distribution, represented as a normal influence diagram (Kenley, 1986, and Shachter and Kenley, 1987). We illustrate the approach with the example of t-PA. In our preliminary trials, this heuristic approach has performed surprisingly well. It is much faster and simpler to use than the numerical integration approach, without much loss in accuracy. Furthermore since dependencies are easy to address, we did not need to make some simplifying assumptions that were made in the earlier analysis of t-PA, and the results of our analysis are in the form of a joint distribution for all of the variables in the model.

Recent news illustrates the need for an accepted approach to address these issues. The Food and Drug Administration advisory panel recommended against the approval of t-PA on May 28, 1987 (Wall Street Journal, 1987). Since no studies have been performed on the long-term survival of patients treated with t-PA, one of the panel members said "I'd like to believe that clot (dissolution) leads to improved survival, but based on the numbers we've seen today, I don't think we can conclude that." The methodology we present provides the mechanism for reaching such a conclusion from the evidence available to the panel.

**Problem Setting**

Medical technology assessments are used to compare alternative therapies for a patients condition based on experimental data. A full analysis considers not just the relative efficacy of different treatments, but also their costs and use of scarce resources. The demand for technology assessments has grown along with escalating health care costs and the efforts to control those costs.

To carry out an assessment properly requires extensive medical and quantitative skills. The data are often incomplete, inconsistent, or noncomparable. The measures of performance might differ from one study to another. Often studies are performed on patients with different characteristics, or there might be variations on the treatments and dosages offered. There are often variations in the manner by which patients enter or leave the study group. All of these complications demand an appreciation of both the medical and statistical implications, and can involve a fair amount of subject "adjustment," to clean up the data for analysis.

As a specific example, we will be considering the efficacy of three treatments for a heart attack: conventional care (CC), and two thrombolytic agents (TA), intravenous streptokinase (IVSK) and t-PA. Our goal is to compute the difference in the probability of one-year mortality given t-PA versus CC. Some of the relevant studies followed patients for a full year after treatment, but others only measure whether reperfusion occurs soon after treatment. In particular, there are no direct data on one-year mortality after t-PA treatment. The influence diagram for a physician treating a patient is given in Figure 1. There is a decision on the Treatment to administer, and subsequent events, Reperfusion, and one-year Mortality. We assume that the probability of Reperfusion depends on the choice of Treatment, and that Mortality depends on both Treatment and Reperfusion.

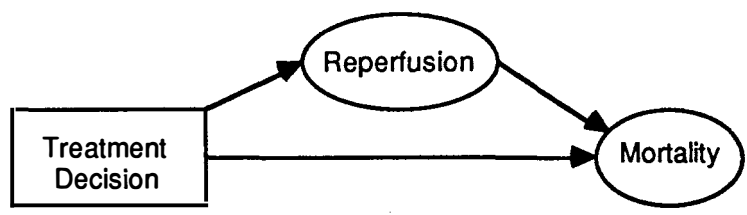

Figure 1. Influence Diagram for the Treatment of a Patient with a Heart Attack.

We have been working for several years to develop a technology assessment processing system for the PC (Eddy, 1986, Hasselblad et al, 1986). The system stores a discrete, but detailed, distribution for each variable (population parameter) in an assessment. It manipulates, adjusts, combines, and displays these distributions. It is menu-driven and relieves the user of the computational burden in performing analyses. This system has been used to perform many assessments, including an analysis of the relative efficacy of t-PA versus conventional care (Eddy, 1987).

There are several weaknesses of the PC system that this new line of research is designed to address. The system in its current design is a powerful calculator, but it has little appreciation of the semantics of the problem. As a result, the user must supply a series of computation commands to perform an analysis. If one of the inputs is revised, the same series of commands must be repeated. Also, there is no way to check the semantic validity of the operations. There are serious and complex issues of evidence evaluation and consolidation, which the user of the system must resolve without help from the system. Finally, the system thinks in terms of one-dimensional uncertainty. It cannot explicitly represent a general joint distribution for two or more dependent variables.

**Normal Influence Diagram**

The key to our new approach is the multivariate normal version of the influence diagram (Howard and Matheson, 1981). All of the manipulations needed to solve problems in a discrete influence diagram (Shachter, 1986) can be easily performed on a Gaussian model as well (Kenley, 1986, Shachter and Kenley, 1987). In fact, this is the simplest possible influence diagram which recognizes dependent, continuous random variables. It assumes that every variable is a linear function of other variables in the model plus some independent, normally-distributed "noise." These assumptions of linearity and normality simplify the representation, assessment, and manipulation of these models. Because the size of the problem grows only with the square of the number of variables, this is a logical model for a problem with many dependent, continuous variables.

The influence diagram representation of a multivariate normal requires a minimum number of parameters to be assessed. For every variable $X_j$ in the model with conditioning variables $C(j)$, we must assess its conditional mean, given that its conditioning variables are at their means,
$$\mu_j = E\{X_j | P_{C(j)} = \mu_{C(j)}\},$$
and its conditional variance, given that its conditioning variables are at their means,
$$v_j = Var\{X_j | X_{C(j)} = \mu_{C(j)}\}.$$
For each conditioning variable $k \in C(j)$, we also need the linear coefficient,
$$b_{kj} = \partial/\partial y_k E\{X_j | X_{C(j)} = x_{C(j)}\},$$
evaluated at $x_{C(j)} = \mu_{C(j)}$.

All of the algorithms have been worked out to manipulate models in this form. They allow one to transform a problem from this representation to the conventional mean vector and covariance matrix form and back again. The different influence diagram operations of expectation and reversal can be simply performed on the problem in the Gaussian influence diagram.

The real advantage of using the influence diagram for the normal model is in the assessment of a multivariate normal model. Unlike the assessment of a covariance matrix, conditional independence simplifies the process and the numbers have a physical interpretation. Maintaining a valid (positive semi-definite) covariance structure is assured by requiring the conditional variances to be nonnegative. Singular (or *almost* singular) covariance matrices are dealt with easily in the influence diagram, whereas they might present problems with more conventional approaches.

## Methodology

In this section we discuss the assumptions made to approximate the assessment model by a normal influence diagram. Every variable in the model, including experimental data, must somehow be transformed into a normal random variable. If the relationships among these transformed variables are nonlinear, then some linear approximation must be used. Our method then interates, to improve this approximation.

The types of variables needed for the analysis of t-PA are probabilities, the difference between probabilities, and the results of randomized controlled trials (RCT). We want a monotone transformation for each variable, such that the errors can all be reasonably considered Gaussian. We used the log-odds transformation for the probabilities, $\log(p/(1-p))$, which has often been used for this purpose. (When a prior distribution was needed for such a variable, the Jeffreys (1961) prior was used for the log-odds, normal with mean 0 and variance $\pi^2$.) A similar transformation, $\log((1+p)/(1-p))$ was used for the difference of probabilities. An RCT transformed variable, Y, with (log-odds of) success probability given by variable X, was assumed to be normally distributed with mean X and variance given by $\zeta(s,2) + \zeta(n-s,2)$ where $\zeta$ is the Riemann zeta function, and s successes were observed out of n trials. This approximation leads to results similar to the beta-binomial conjugate distribution model, after the variable is instantiated with a value of $\psi(s)-\psi(n-s)$, where $\psi$ is the Euler psi or digamma function.

The linear approximation to nonlinear relationships were obtained by a simple heuristic. There were two functions for which it was needed. First, it was used to define the difference between probability of mortality for t-PA and CC. While the difference would be linear if the variables were the probabilities, it is nonlinear in terms of the log-odds transformations. The other function was the chaining relationship, expressing the probability of mortality in terms of probabilities of reperfusion and mortality given reperfusion or not reperfusion:

Pr{ Mortality } = Pr { Mortality | Reperfusion } · Pr { Reperfusion }
+ Pr { Mortality | Not Reperfusion } · (1 − Pr { Reperfusion }).

To approximate the expected value of this function, it was evaluated at the expected value of its arguments, which were themselves approximated by the inverse transforms of their means. The gradient of this function was evaluated at the same point.

This linear approximation is obviously a weak point in the model. However, its accuracy was improved through iteration. The linear approximation in the first iteration used the prior distribution for each mean (before instantiation). After each iteration the linear approximation was recalculated using the gradient at the posterior mean instead of the prior. In this and other examples, this technique has converged quickly.

233

## Example of t-PA Analysis with Heuristic Method

The t-PA assessment is represented by the influence diagram in Figure 2. In the figure, the RCT data to be instantiated are indicated by shading. These data are given in the table below:

| Study | Treatment | Outcome | Successes | Trials |
|---|---|---|---|---|
| TIMI (1985) | t-PA | Reperfusion | 78 | 118 |
|  | IVSK | Reperfusion | 44 | 122 |
| Collen et al (1984) | t-PA | Reperfusion | 25 | 33 |
|  | CC | Reperfusion | 25 | 33 |
| Yusuf et al (1985) | IVSK | Mortality | 412 | 2672 |
|  | CC | Mortality | 501 | 2612 |
| Kennedy et al (1985) | TA | Mortality | Reperfusion | 5 | 93 |
|  | TA | Mortality | No Reperfusion | 6 | 41 |
|  | CC | Mortality | Reperfusion | 0 | 14 |
|  | CC | Mortality | No Reperfusion | 17 | 102 |

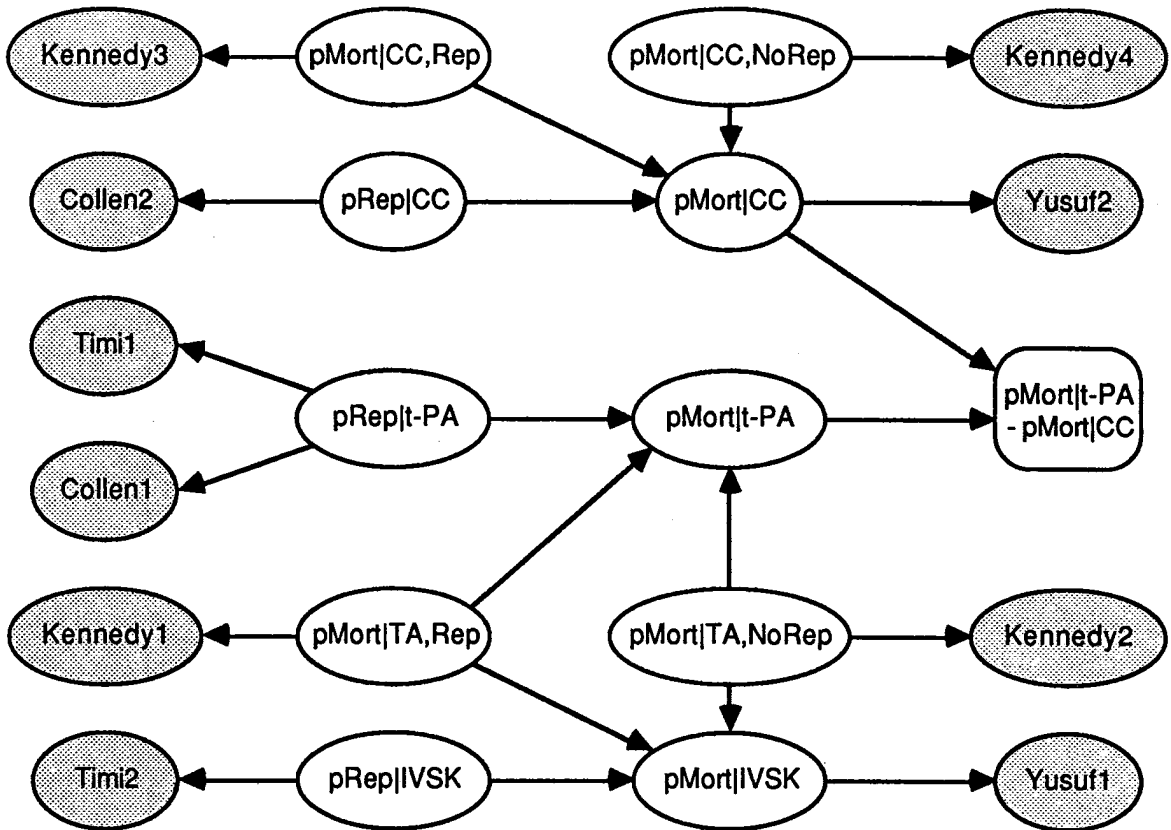

*Figure 2. Influence Diagram for the Analysis of t-PA. The experiments to be instantiated are shaded.*

A simple test system was written for this problem using ExperCommonLisp on the Macintosh computer. It takes one simple function call for each variable defined in the model. The first six iterations from the test system are:

| Pr{Mort.\|t-PA}–Pr{Mort.\|CC} | Pr{Mort.\|CC} | Pr{Mort.\|t-PA} | Pr{Mort.\|IVSK} |
|---|---|---|---|
| -0.070061 | 0.1916976 | 0.1519070 | 0.1519070 |
| -0.088682 | 0.1911351 | 0.1087691 | 0.1530587 |
| -0.087049 | 0.1913679 | 0.1044005 | 0.1530891 |
| -0.087141 | 0.1913791 | 0.1042373 | 0.1530686 |
| -0.087146 | 0.1913792 | 0.1042328 | 0.1530680 |
| -0.087147 | 0.1913797 | 0.1042325 | 0.1530680 |

This problem takes minutes to enter into the computer and takes under two minutes to solve. The more polished and friendly PC version takes a couple of hours to enter the problem and solve it. The answer from that system is that the difference has a mean of –8% (a decrease in the probability of mortality of 8%) with a standard deviation of about 2.5%. This compares well with the quick and easy approximation of -8.7% with a standard deviation of 1.7%.

## Conclusions

Much work remains to be done with this heuristic approach. The techniques need to be extended to include other forms of data and relationships found in medical technology assessment. These include:
- other types of experiments, such as retrospective case control trials,
- models of systematic errors in the observation of outcomes, or dilution and contamination (patients refusing or seeking treatment despite their assignment to a treatment or control group), and
- adjustments, to account for differences between studies, such as dosages and patient characteristics.

We also need to refine and validate the approximations taken. In particular, we will be working on the step in which the expected value of a transformed variable is approximated by a linear function of other transformed variables. The validation work can proceed in two directions: comparing the results of this model with other assessments performed with the PC system and using other approaches to approximate non-Gaussian multivariate models.

An interesting extension of this approach would be the development of an expert system interface to build the influence diagram model. Such a system would need a knowledge base about the construction of these types of statistical models. If successful, it would further reduce the statistical burden on the medical researcher, and permit medical analysts who are less mathematically sophisticated to make full use of the system.
   Finally, the results and techniques in this paper extend beyond medical technology assessment to many problems in which experimental results are being integrated to form probability distributions for population parameters. The problems we have had to overcome in organizing and analyzing our assessments are endemic to all such learning systems. The task of gathering, interpreting, and distilling the results from experimental studies is the one task that can never be automated. Our goal is to provide an expert system environment to complete the analysis once those data have been prepared.